\documentclass{article}
\usepackage{amssymb}
\usepackage{ijcai25}

\usepackage{times}
\usepackage{wrapfig}
\usepackage{soul}
\usepackage{url}
\usepackage[hidelinks]{hyperref}
\usepackage[utf8]{inputenc}
\usepackage[small]{caption}
\usepackage{graphicx}
\usepackage{amsmath}
\usepackage{amsthm}
\usepackage{booktabs}
\usepackage{algorithm}
\usepackage{algorithmic}
\usepackage[switch]{lineno}
\usepackage{enumitem}
\usepackage{xcolor}
\usepackage{subcaption}

\title{Uncertainty-aware Predict-Then-Optimize Framework for \\Equitable Post-Disaster Power Restoration}

\author{
Lin Jiang\textsuperscript{1},
Dahai Yu\textsuperscript{1},
Rongchao Xu\textsuperscript{1},
Tian Tang\textsuperscript{2},
Guang Wang\textsuperscript{1}\thanks{Corresponding author}\\
\affiliations
\textsuperscript{1}Department of Computer Science, Florida State University\\
\textsuperscript{2}Askew School of Public Administration and Policy, Florida State University\\
\emails
\{lj23d, dahai.yu, rx21a, ttang4\}@fsu.edu, guang@cs.fsu.edu
}

\begin{document}

\maketitle

\begin{abstract}
The increasing frequency of extreme weather events, such as hurricanes, highlights the urgent need for efficient and equitable power system restoration. Many electricity providers make restoration decisions primarily based on the volume of power restoration requests from each region. However, our data-driven analysis reveals significant disparities in request submission volume, as disadvantaged communities tend to submit fewer restoration requests. This disparity makes the current restoration solution inequitable, leaving these communities vulnerable to extended power outages. To address this, we aim to propose an equity-aware power restoration strategy that balances both restoration efficiency and equity across communities. However, achieving this goal is challenging for two reasons: the difficulty of predicting repair durations under dataset heteroscedasticity, and the tendency of reinforcement learning agents to favor low-uncertainty actions, which potentially undermine equity. To overcome these challenges, we design a predict-then-optimize framework called EPOPR with two key components: (1) Equity-Conformalized Quantile Regression for uncertainty-aware repair duration prediction, and (2) Spatial-Temporal Attentional RL that adapts to varying uncertainty levels across regions for equitable decision-making. Experimental results show that our EPOPR effectively reduces the average power outage duration by 3.60\% and decreases inequity between different communities by 14.19\% compared to state-of-the-art baselines.

\end{abstract}

\footnotetext[1]{Accepted to the IJCAI 2025 (AI and Social Good Track).}

\vspace{-15pt}
\section{Introduction}
Power restoration following extreme weather events is crucial for social well-being. In the Southeastern United States, many areas are particularly vulnerable to hurricanes, which can cause severe disruptions. For instance, in September 2024, Category 5 Hurricane Helene struck Florida, Georgia, and North Carolina, affecting 1.7 million people and causing an estimated \$78.7 billion in damages \cite{heleneUSA}. Through interviews with government agencies and real-world data analysis in Florida, we found that obtaining real-time power status across all communities and households after a hurricane remains a significant challenge. As a result, the government has launched online platforms, such as Florida's 311 system \cite{xu2020closing}, to allow residents to submit power repair requests. These submissions help authorities prioritize regional power restoration, often giving precedence to communities reporting the highest number of outages.

Although the current strategy can be effective in certain cases, it often exacerbates existing inequalities by disproportionately impacting economically and socially disadvantaged communities. Due to limited awareness of submission channels, residents in these communities typically file fewer requests, making them more likely to be overlooked in government decision-making. Therefore, it is crucial to develop an equity-aware power restoration strategy that balances restoration efficiency with fairness across all communities.

However, two major challenges arise in achieving this.  
First, heteroscedasticity in the training data \cite{white1980heteroskedasticity} poses a challenge to regional repair duration prediction. Repair durations, which are critical for determining restoration sequences, are typically derived from historical repair request data. However, as our findings suggest, the volume of repair requests varies significantly across regions, introducing heteroscedasticity into the data. This variability hampers the ability to make deterministic predictions, and directly applying traditional uncertainty quantification methods such as conformal prediction \cite{shafer2008tutorial} can result in inequities across sensitive features.
Second, traditional optimization methods like reinforcement learning (RL) prioritize information with lower uncertainty to minimize error accumulation \cite{kumar2020conservative}. Hence, directly applying these methods would lead to regions with smaller local variances being prioritized, which conflicts with our equity objective.

To address these challenges, we propose an \underline{E}quity-aware \underline{P}redict-then-\underline{O}ptimize \underline{P}ower \underline{R}estoration framework, called \textbf{EPOPR}. The objective is to minimize the total outage duration while ensuring that outage durations across communities satisfy equity criteria. EPOPR consists of two key components:  
(1) \textbf{Equity-Conformalized Quantile Regression (ECQR)} for uncertainty-aware repair duration prediction. In a heteroscedastic dataset, some sensitive features may have limited data coverage and wider prediction intervals, potentially compromising equity in subsequent decisions. To address this, ECQR incorporates equity-based uncertainty calibration to maintain uniform average coverage across sensitive features. Additionally, it employs dynamic prediction intervals that adapt to varying levels of dispersion across sensitive features.  
(2) \textbf{Spatial-Temporal Attentional Soft Actor-Critic (STA-SAC)} for optimizing the repair sequence. The key innovation lies in the spatial-temporal attention-based Actor, which captures spatiotemporal dependencies among uncertainties while handling the dynamic nature of the action set. Meanwhile, we integrate the Lagrange Multiplier \cite{bertsekas2014constrained} into Soft Actor-Critic \cite{haarnoja2018soft}, our base RL model, to solve the multi-objective optimization problem that balances both efficiency and equity.

The key contributions of this paper include:
\begin{itemize}
    \item From a data-driven perspective, our analysis of real-world power restoration datasets reveals two key insights: (1) The current power restoration process is inequitable, as it tends to prioritize communities with higher volumes of repair requests, typically those that are economically and socially advantaged. (2) Repair request submissions vary significantly across regions, introducing heteroscedasticity into the training data and complicating subsequent decision-making.

    \item From a technical design perspective, inspired by our data-driven findings, we propose an equity-aware predict-then-optimize power restoration framework, EPOPR, with novel enhancements at both the prediction and decision-making stages. For repair duration prediction, we introduce ECQR, which incorporates equity-aware uncertainty calibration to ensure uniform average coverage across sensitive features, while employing dynamic prediction intervals to preserve overall statistical efficiency. For repair sequence decision-making, we develop STA-SAC, an RL method designed to minimize total outage duration while enforcing equity-aware constraints. Specifically, STA-SAC employs a Spatial-Temporal Attentional Actor to jointly capture predictive uncertainty and its spatiotemporal dependencies.

    \item We comprehensively evaluate EPOPR using the real-world power outage datasets from Tallahassee, Florida. Experimental results demonstrate that EPOPR reduces average outage duration by 3.60\% and decreases inequity among regions by 14.19\%, outperforming the best baseline. Specifically, we independently assess our prediction method, ECQR, which enhances prediction performance in disadvantaged regions and produces a more equitable output. This equity-aware prediction further enables our decision-making method STA-SAC to significantly narrow the disparity in power outage durations across regions with different income levels.

\end{itemize}

\section{Data Analysis And Motivation}
\begin{figure}[htbp]\centering
\begin{minipage}[htbp]{0.50\linewidth}
    \includegraphics[width = \linewidth,keepaspectratio=true]{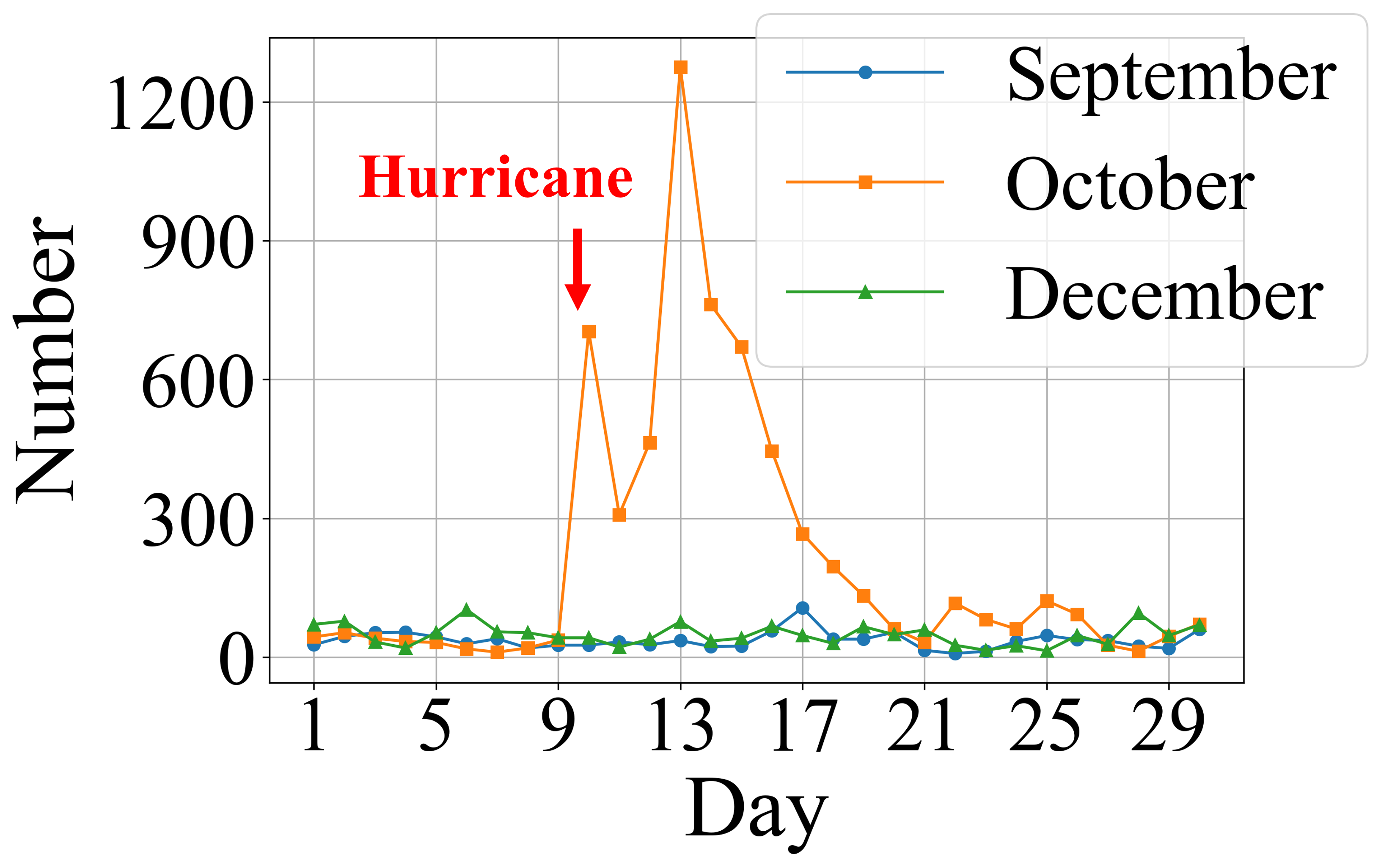}
    \captionsetup{font={small}}
    \caption{Repair Requests in Different Days}
    \label{fig:request temporal}
\end{minipage}
\hspace{2mm}
\begin{minipage}[htbp]{0.44\linewidth}
    \includegraphics[width = \linewidth,keepaspectratio=true]{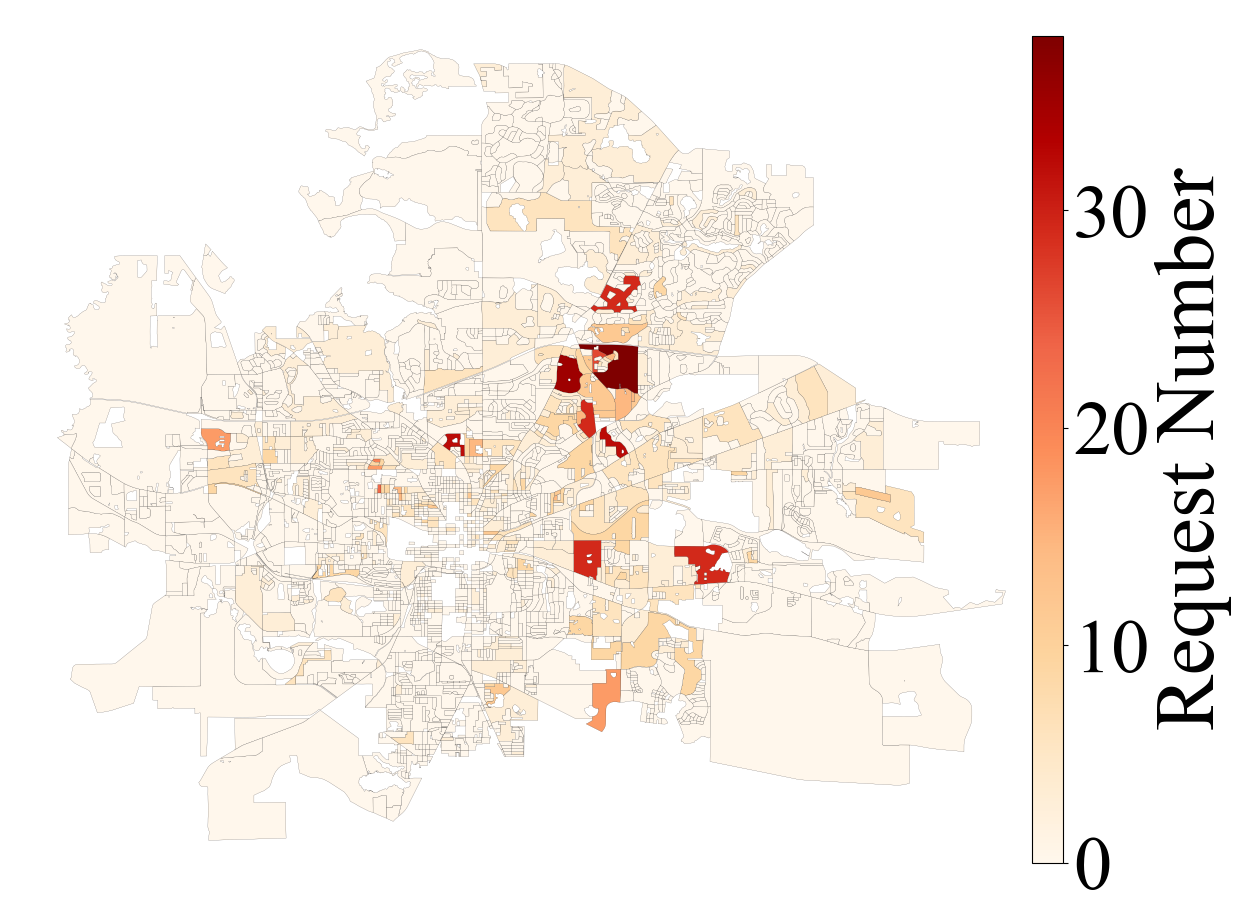}
    \captionsetup{font={small}}
    \caption{Post-Hurricane Repair Requests across Regions}
    \label{fig:request spatial}
\end{minipage}
\end{figure}

In this project, we collaborate with the \textbf{City of Tallahassee Government} in Florida to improve its power outage restoration services. The city provided us with detailed household-level datasets on outage repair requests and electricity usage after signing a non-disclosure agreement. Both datasets contain one year of data records from Tallahassee in 2018. Notably, they capture the city's power restoration process during \textbf{Hurricane Michael in October 2018}—a Category 5 storm that caused large-scale outages in the city \cite{nhcMichael2019}. Moreover, our collaborator from \textbf{Public Administration and Policy} offered detailed investigations and analyses from a \textbf{Social Science} perspective \cite{xu2020closing}, helping us identify key variables influencing government decision-making and understand how these factors may contribute to inequities in the power restoration process. Building on these collaborations, we conducted an in-depth, data-driven analysis that yielded the following key findings:

(i) The number of repair requests surges significantly after a hurricane, yet their distribution varies considerably across regions. Figures \ref{fig:request temporal} and \ref{fig:request spatial} illustrate the fluctuations in repair request submissions across both spatial and temporal dimensions. In the temporal dimension, we observe a sharp increase in repair requests within the first week after the hurricane, indicating that residents rely heavily on this method to report power outages.
In the spatial dimension, the number of requests varies significantly across regions due to multiple factors, including the severity of outages and residents' awareness of available reporting channels.

% \begin{figure}[htbp]
%     \centering
%     \begin{subfigure}[t]{0.49\linewidth}
%         \includegraphics[width = \linewidth, keepaspectratio=true]{fig/request temporal.png}
%         \caption{Repair Requests in Different Days}
%         \label{fig:request temporal}
%     \end{subfigure}
%     \hspace{0.04\linewidth}
%     \begin{subfigure}[t]{0.44\linewidth}
%         \includegraphics[width = \linewidth, keepaspectratio=true]{fig/request spatial.png}
%         \caption{Post-Hurricane Repair Requests across Regions}
%         \label{fig:request spatial}
%     \end{subfigure}
% \end{figure}

\begin{figure}[b]\centering
\begin{minipage}[htbp]{0.43\linewidth}
    \includegraphics[width = \linewidth,keepaspectratio=true]{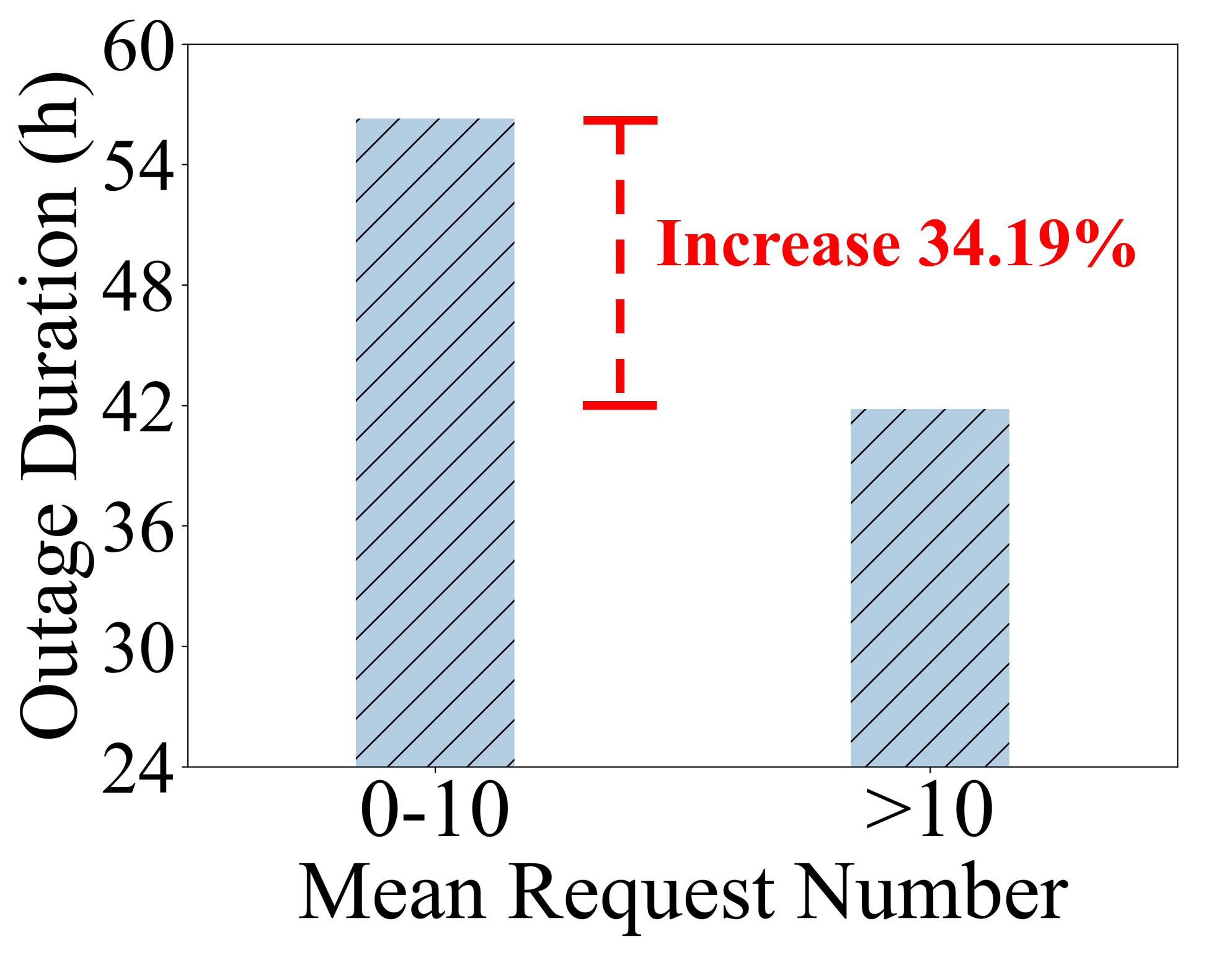}
    \captionsetup{font={small}}
    \caption{Request Number vs. Outage Duration}
    \label{fig:request number vs outage duration}
\end{minipage}
\hspace{2mm}
\begin{minipage}[htbp]{0.51\linewidth}
    \includegraphics[width = \linewidth,keepaspectratio=true]{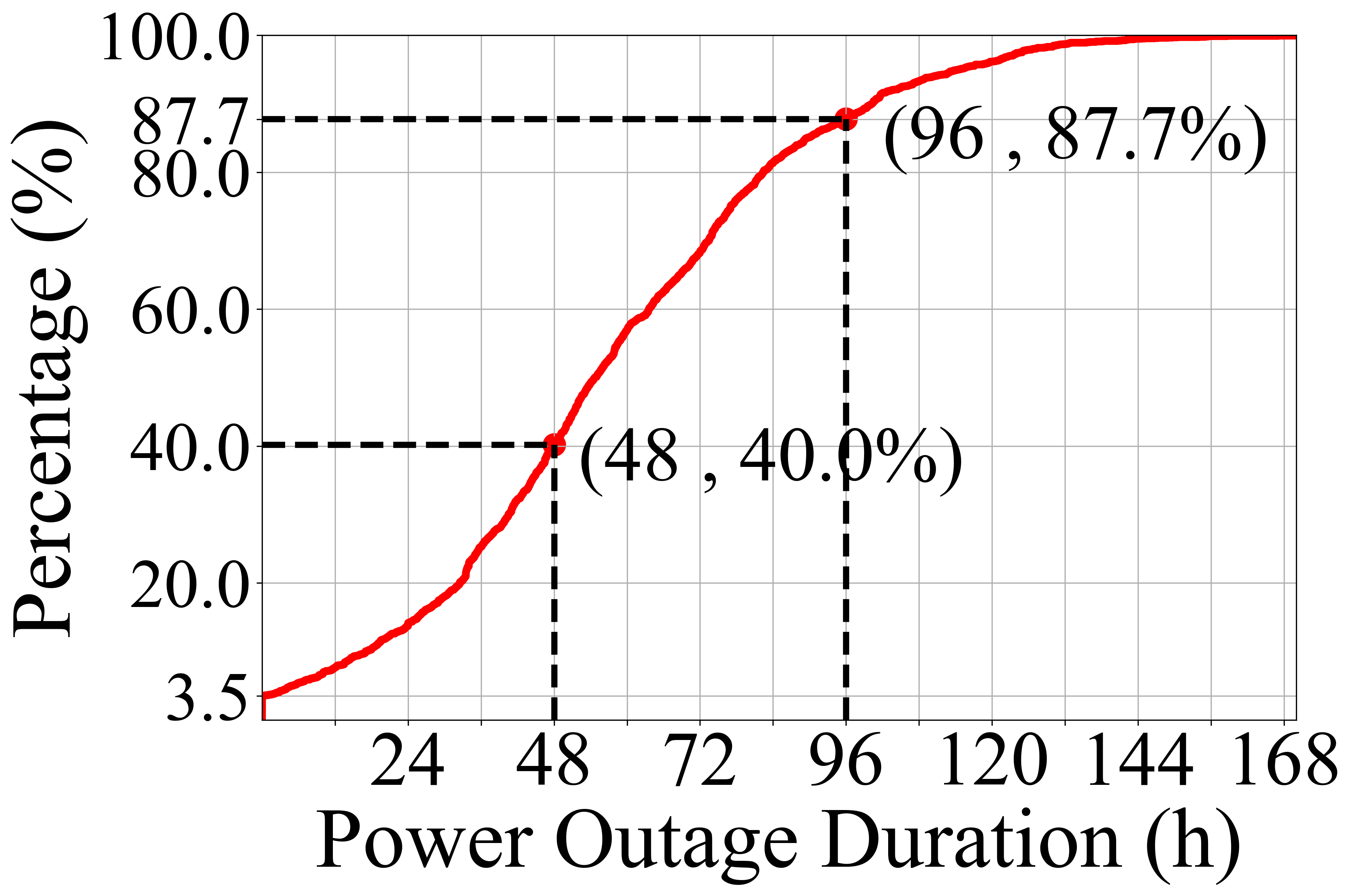}
    \captionsetup{font={small}}
    \caption{Power Outage Duration during Hurricane}
    \label{fig:power outage duration}
\end{minipage}
\end{figure}

% \begin{figure}[b]
%     \centering
%     \begin{subfigure}[t]{0.44\linewidth}
%         \includegraphics[width=\linewidth, keepaspectratio=true]{fig/Request Number v.s. Outage Duration.png}
%         \captionsetup{font={small}}
%         \caption{Request Number vs. Outage Duration}
%     \label{fig:request_number_vs_outage_duration}
%     \end{subfigure}
%     \hspace{0.04\linewidth}
%     \begin{subfigure}[t]{0.5\linewidth}
%         \includegraphics[width=\linewidth, keepaspectratio=true]{fig/Power Outage Duration.png}
%         \captionsetup{font={small}}
%         \caption{Power Outage Duration during Hurricane}
%         \label{fig:power_outage_duration}
%     \end{subfigure}
% \end{figure}

\begin{figure}[t]\centering
\begin{minipage}[htbp]{0.48\linewidth}
    \includegraphics[width = \linewidth,keepaspectratio=true]{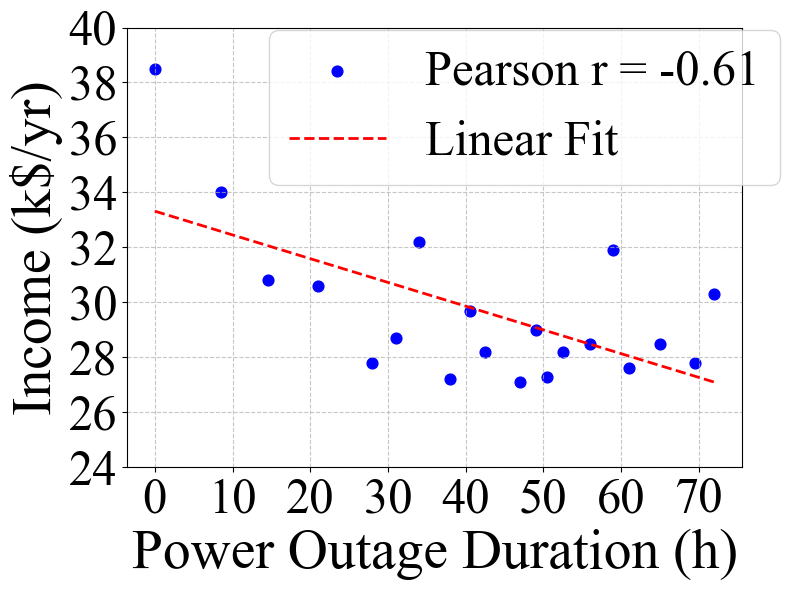}
    \captionsetup{font={small}}
    \caption{Income vs. Power Outage Duration}
    \label{fig:Income}
\end{minipage}
\hspace{2mm}
\begin{minipage}[htbp]{0.45\linewidth}
    \includegraphics[width = \linewidth,keepaspectratio=true]{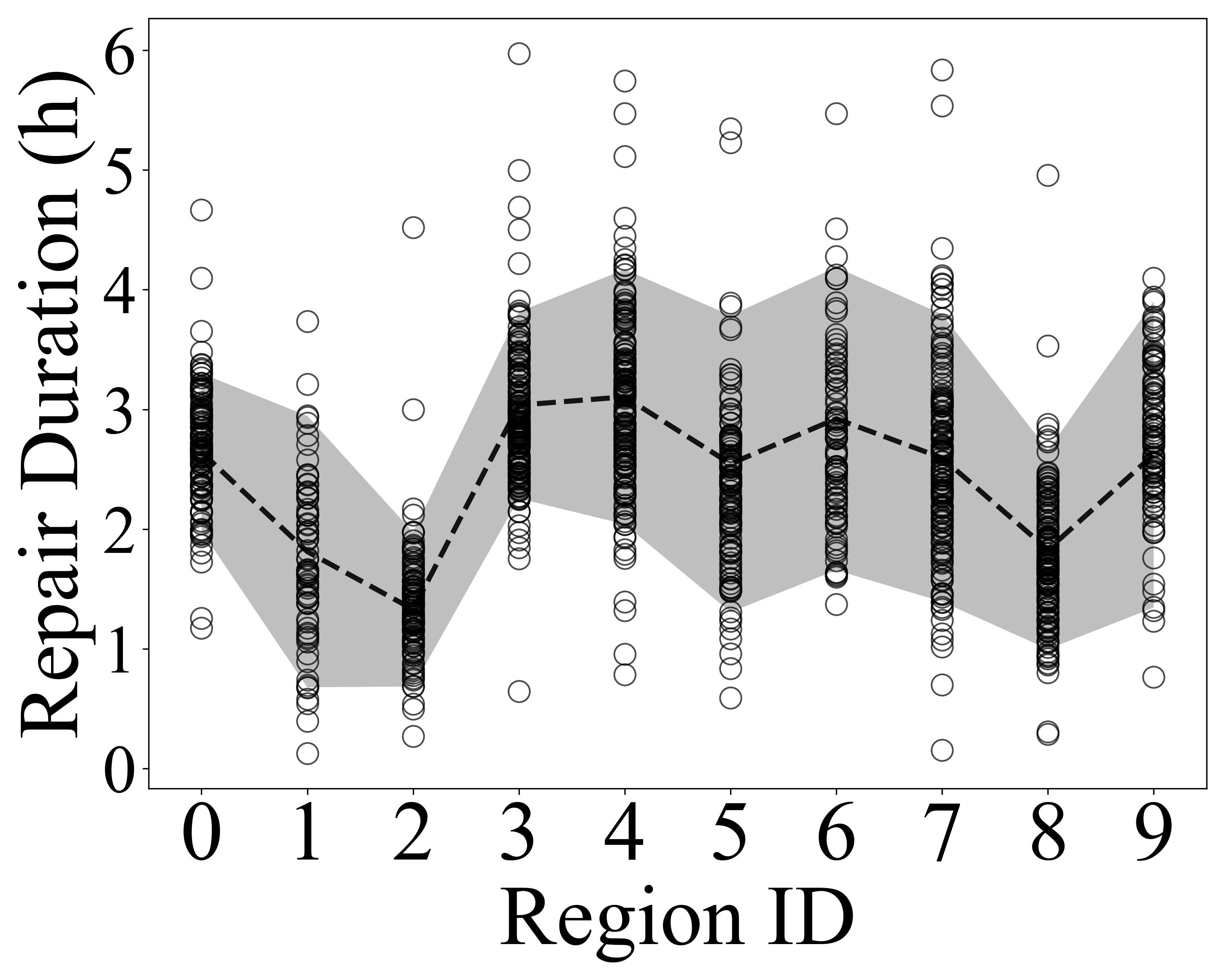}
    \captionsetup{font={small}}
    \caption{Repair Duration in Different Regions}
    \label{fig:Repair Duration}
\end{minipage}
\end{figure}

(ii) The current power restoration strategy in Tallahassee disproportionately disadvantages economically and socially disadvantaged communities. Interviews with government officials revealed that obtaining real-time power status across all regions after a hurricane remains a significant challenge. Consequently, the government often relies on repair request volume to determine the restoration sequence. However, this approach leads to inequities, as regions with fewer repair requests are more likely to experience prolonged power outages. As shown in Figure \ref{fig:request number vs outage duration}, regions that submitted fewer repair requests (10 or fewer) experienced 34.19\% longer power outage durations than those with over 10 requests. In Figure \ref{fig:power outage duration}, we present the cumulative distribution function (CDF) of power outage durations across all regions during the hurricane. The data reveals that 40\% of outages were restored within two days, while over 12\% of residents still experienced outages lasting more than four days. We also examine the socioeconomic status of regions experiencing longer power outages. Using average annual income as a representative metric, Figure \ref{fig:Income} reveals a negative correlation between power outage duration and income levels, with a Pearson correlation coefficient of -0.61. This finding indicates that lower-income regions tend to endure longer power outages. Motivated by this, our research aims to develop an AI-driven power restoration strategy that enhances social equity.

(iii) The historical regional repair request data exhibits heteroscedasticity. In Figure \ref{fig:Repair Duration}, we visualize the distribution of historical repair durations across different regions. The shaded area represents the 90\% quantile interval (ranging from the 5th to the 95th percentile), while the dashed line indicates the midpoint of this range. It can be observed that the variance of historical repair durations differs significantly across regions. For example, Region 2 has a narrower quantile interval, whereas Regions 4 and 7 have wider intervals. This pronounced heteroscedasticity poses challenges for applying deterministic prediction methods. Moreover, traditional uncertainty-aware prediction approaches often introduce bias, as their predictive accuracy tends to be higher for groups with more data (e.g., richer communities), further exacerbating disparities across socioeconomic groups.

% \begin{figure}[t]
%     \centering
%     \begin{subfigure}[t]{0.48\linewidth}
%         \includegraphics[width=\linewidth, keepaspectratio=true]{fig/Income.png}
%         \captionsetup{font={small}}
%         \caption{Income vs. Power Outage Duration}
%         \label{fig:Income}
%     \end{subfigure}
%     \hspace{0.04\linewidth}
%     \begin{subfigure}[t]{0.45\linewidth}
%         \includegraphics[width=\linewidth, keepaspectratio=true]{fig/Repair Duration.png}
%         \captionsetup{font={small}}
%         \caption{Repair Duration in Different Regions}
%         \label{fig:Repair Duration}
%     \end{subfigure}

% \end{figure}

\section{Methodology}
\subsection{ System Overview}
In this paper, we design an equity-aware predict-then-optimize power restoration framework called EPOPR, considering both recovery efficiency and equity. The problem is formulated as follows:
\begin{gather}
    \min \sum_{ k \in \{0, 1, 2\}} P(F_k) \mathbb{E}[T_{outage} \mid F_k] \label{eq:1}\\
    \text{s.t.} \; \max_{k_1, k_2} W(P(T_{outage} \mid F_{k_1}), P(T_{outage} \mid F_{k_2})) \leq d \label{eq:2}
\end{gather}

In Equation \ref{eq:1}, we present our \textbf{optimization objective}, which seeks to minimize the average expected outage duration across all regions. The feature $F$ represents an equity-related feature. Taking the average income as an example, we categorize $F$ into three groups: high, middle, and low.

In Equation \ref{eq:2}, we present our \textbf{equity-aware constraints}. Inspired by the fairness definition of demographic equity \cite{pessach2022review}, we define equity as the condition that the difference in outage duration distributions across different sensitive groups of regions does not exceed a predefined upper bound $d$—for example, the difference in outage duration between high-income and low-income regions should not be too large. To better capture the overall distributional difference, we use the Wasserstein distance $W(P(T_{\text{outage}} \mid F_{k_1}), P(T_{\text{outage}} \mid F_{k_2}))$ to represent the disparity \cite{panaretos2019statistical}.

According to our optimization formulation, the regional outage duration $T_{outage}(F_k)$ (abbreviated as $T_{outage}$ when the sensitive feature $F_k$ is not explicitly considered) plays a crucial role. The outage duration for region $m$, $T^{(m)}_{outage}$, consists of two components: waiting duration $T^{(m)}_{waiting}$ and repair duration $T^{(m)}_{repair}$. The waiting duration, $T^{(m)}_{waiting}$, can further be represented as the sum of the repair durations of all regions preceding region $m$, i.e., $\sum_{j=1}^{m-1} T^{(j)}_{repair}$, and the time spent on the road, $T_{travel}$.
The equation is shown below:
\begin{align}
    T^{(m)}_{outage} &= T^{(m)}_{waiting} + T^{(m)}_{repair} \nonumber \\
    &= \textstyle \sum_{j=1}^{m-1} T^{(j)}_{repair} + T_{travel} + T^{(m)}_{repair} \label{eq:3}
\end{align}

\begin{figure}[htbp]  \centering
    \centering
    \includegraphics[width=1\linewidth, keepaspectratio=true]{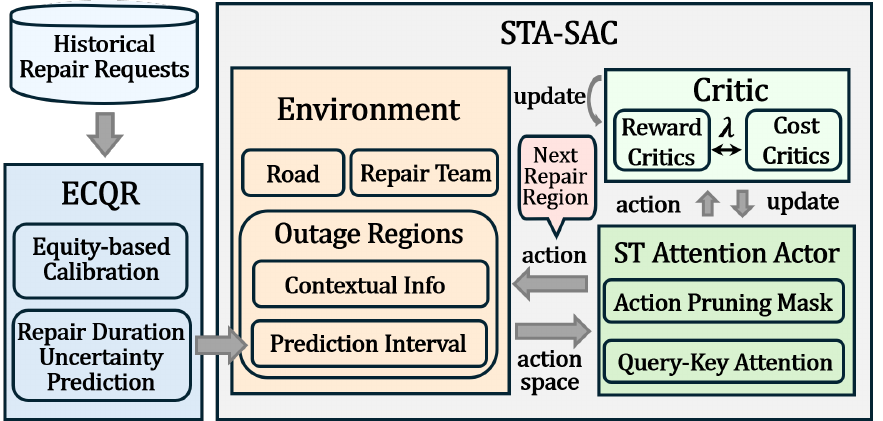}
    \caption{The Framework of EPOPR}
    \label{fig:EPOPR}
\end{figure}

Based on the formulation in Equation \ref{eq:3}, we break down our optimization problem into two key subproblems: predicting the outage repair duration for each region, which relates to $T^{(m)}_{repair}$, and determining the repair sequence, which is related to $T^{(m)}_{waiting}$. Accordingly, our EPOPR framework consists of two main components: a prediction module called ECQR that estimates the repair duration $T_{repair}$ for each region, and a decision module called STA-SAC that determines the repair sequence across different regions. An overview of the EPOPR framework is illustrated in Figure \ref{fig:EPOPR}.

 \subsection{ECQR for Prediction}
In this section, we describe how we estimate repair durations \( T_{\text{repair}} \) across different regions. A key challenge, as discussed earlier, is the substantial variation in repair request volumes across regions. This leads to an unevenly distributed training set and results in heteroscedasticity. For such high-variance data, deterministic predictions often underperform, especially in underrepresented regions. To address this, we adopt uncertainty-aware prediction. However, this approach alone may still lead to unfair outcomes. Due to heteroscedasticity, the variance in repair durations differs across regions (as shown in Figure~\ref{fig:Repair Duration}), and regions with higher variance often experience poorer prediction performance, making it more difficult for them to receive equitable restoration services.

To handle this, we propose Equity-Conformalized Quantile Regression (ECQR) based on traditional CQR \cite{romano2019conformalized}. Our method has two distinct advantages: (1) We provide a group-based calibration factor to ensure equitable coverage. Although this may slightly reduce prediction coverage for advantaged groups, it significantly narrows the coverage gap between advantaged and disadvantaged groups. (2) Inheriting the advantages of original CQR, ECQR also provides dynamic prediction intervals for different regions, making it particularly suitable for heteroscedastic datasets and improving overall prediction performance.

We now introduce ECQR in detail. Let the original dataset \( D \) be denoted as \( \{(x_i, f_i, y_i): i \in D \} \), where \( x_i \) represents the features of sample \( i \), \( f_i \) corresponds to equity-related sensitive features, and \( y_i \) denotes the target variable. In our scenario, \( x_i \) is a 9-dimensional vector including the region's location and contextual information, and \( f_i \) represents the region's average annual income, divided into three sensitive groups, with \( y_i \) denoting the region's repair duration.

The dataset \( D \) is split into a training dataset \( D_1 \) and a calibration dataset \( D_2 \). First, we use Quantile Regression Forest (QRF) \cite{meinshausen2006quantile} to fit two conditional quantile functions \( \hat{q}^{\alpha}_{lo} \) and \( \hat{q}^{\alpha}_{hi} \), as shown below:
\begin{equation}
    \left\{\hat{q}_{\alpha_{lo}}, \hat{q}_{\alpha_{hi}}\right\} \leftarrow QRF\left(\left\{(x_i, y_i) : i \in D_1\right\}\right) \label{eq:4}
\end{equation}
Where \(\hat{q}_{\alpha_{lo}}\) and \(\hat{q}_{\alpha_{hi}}\) represent the conditional quantile functions for the lower and upper bounds of prediction \(\alpha\), respectively. The definition for $\hat{q}_{\alpha}(x)$ is shown as below:
\begin{gather}
F(y \mid X = x) := P\{Y \leq y \mid X = x\} \label{eq:5} \\
\hat{q}_\alpha(x) := \inf\{y \in \mathbb{R} : F(y \mid X = x) \geq \alpha\} \label{eq:6}
\end{gather}

For example, if we pursue the prediction interval with coverage rate \( \alpha \) as 90\%, then \( \alpha_{lo} = \frac{1-\alpha}{2} \) and \( \alpha_{hi} = \frac{1+\alpha}{2} \). We use $PI_{QRF}(x_{i})$ to represent the prediction interval for sample $i$:
\begin{equation}
    PI_{QRF}(x_{i}) = [\hat{q}_{\alpha_{lo}}(x_{i}), \hat{q}_{\alpha_{hi}}(x_{i})] \label{eq:7}
\end{equation}

Directly applying the prediction interval \( PI_{QRF} \) learned from quantile regression is impractical because the validity of estimated intervals is only guaranteed for specific models rather than our heteroscedasticity setting \cite{takeuchi2006nonparametric}. Therefore, we compute conformity scores in the calibration dataset \( D_2 \) to quantify prediction interval error and calibrate it. The conformity score \( E \) is calculated as follows:
\begin{equation}
    E_i := \max\left\{ \hat{q}_{\alpha_{lo}}(x_i) - y_i, y_i - \hat{q}_{\alpha_{hi}}(x_i) \right\}, \quad i \in D_2 \label{eq:8}
\end{equation}

In original CQR, the method sorts all \( E_i \) values and selects a specific quantile of \( E_i \) to calibrate the prediction interval:  
\begin{equation}
    E_{(1)} \leq E_{(2)} \leq \dots \leq E_{(|D_2|)} \label{eq:9}
\end{equation}
Specifically, the method selects the \( \lceil \alpha(|D_2|+1) \rceil \)-th smallest conformity score \( E_{(\lceil \alpha(|D_2|+1) \rceil)} \) as the calibration factor $Q_\alpha(E,D_2)$ for adjusting the prediction interval. Where $\lceil \rceil$ represents the ceiling function. The prediction interval $PI_{CQR}(x_{i})$ can be represented as:
\begin{equation}
\begin{aligned}
PI_{CQR}(x_{i}) = 
\big[ & \hat{q}_{\alpha_{lo}} (x_{i}) - Q_{\alpha}(E, D_2), \\
      & \hat{q}_{\alpha_{hi}} (x_{i}) + Q_{\alpha}(E, D_2) \big] \label{eq:10}
\end{aligned}
\end{equation}

However, the original CQR does not account for the influence of the sensitive feature \( F \). As shown in Equation~\ref{eq:9}, the sorting of all conformity scores \( E_i \) is independent of the corresponding sensitive feature \( f_i \). In other words, all samples in the calibration set \( D_2 \) are sorted without considering their sensitive attributes. As a result, the scores covered by the calibration factor \( Q_\alpha(E, D_2) \) (i.e., from \( E_1 \) to \( E_{\lceil \alpha (|D_2|+1) \rceil} \)) may have imbalanced sensitive feature distributions. For instance, due to higher variance, samples from low-income regions are less likely to fall within the calibrated range.

In ECQR, we redefine the calibration dataset $D_2$ based on the sensitive feature \( F \):
\begin{equation}
    D_2(F_k) = \{i : i \in D_2 \text{ and } f_i = F_k\} \label{eq:11}
\end{equation}
Where $F_k$ means the $k$-th sensitive group in sensitive feature $F$. 
After each sensitive group $F_k$ has a corresponding calibration dataset $D_2(F_k)$, we calculate the group-based calibration factor $Q_\alpha(E,D_2(F_k))$ separately through the $D_2(F_k)$ rather than the whole $D_2$. Based on that, we get the prediction interval $PI_{ECQR}(X_{i},f_{i})$ for our method $ECQR$:
\textbf{\begin{equation}
\small
\begin{aligned}
PI_{ECQR}(x_{i},f_{i})  = 
& \big[\hat{q}_{\alpha_{lo}} (x_{i},f_{i}) - Q_{\alpha}(E, D_2(F_k)), \\
      & \hat{q}_{\alpha_{hi}} (x_{i},f_{i}) + Q_{\alpha}(E, D_2(F_k)) \big] \label{eq:12}
\end{aligned}
\end{equation}}
From Equation \ref{eq:12}, it can be observed that when predicting the prediction interval \( PI_{ECQR}(x_{i},f_{i}) \) for sample \( i \), we account for its sensitive feature \( f_{i} \) and apply the corresponding calibration factor \( Q_{\alpha}(E, D_2(F_k)) \) based on the sensitive group \( F_k \) associated with the sample. This group-based approach ensures that the system's output maintains the same theoretical coverage across all sensitive groups.

To conclude, our ECQR represents uncertainty prediction outputs as prediction intervals $PI_{ECQR}(x_{i},f_{i})$, offering two key advantages: equitable coverage across all sensitive groups and the dynamic prediction interval for every sample $i$. In the next section, we will introduce the decision-making module and explain how the prediction intervals are used in STA-SAC for equity-aware power restoration.

 \subsection{STA-SAC for Decision-Making}

 \subsubsection{CMDP Problem Formulation}
 Formally, we model our equity-aware power restoration problem as a Constrained Markov Decision Process (CMDP), and propose a new reinforcement learning algorithm called STA-SAC to solve it. We define the CMDP problem $\mathcal{G}$ as the 6-tuple: $\mathcal{G} = \{\mathcal{S},\mathcal{A},\mathcal{P},\mathcal{R},\mathcal{C},\mu\}$,
where $\mathcal{S}$ is the state space, $\mathcal{A}$ is the action space, $\mathcal{P}: \mathcal{S} \times \mathcal{A} \times \mathcal{S} \rightarrow [0,1]$ denotes the transition probability function, $\mathcal{R}$ represents the reward function, $\mathcal{C}$ represents the cost function, $\mu$ is the initial state distribution. The details of the CMDP $\mathcal{G}$ in our problem are shown below.
\begin{itemize}[leftmargin=*]
    \item \textbf{State $\mathcal{S}$}: We consider the power repair team as the agent. We define the state \( s_t \) of the agent at step \( t \) as \( s_t = ST_t \), where \( ST_t \) represents the agent’s spatiotemporal state, encompassing the current region \( r \), the current time \( T_t \), and its current coordinates.
    \item \textbf{Action $\mathcal{A}$}: We define the action as selecting the next power outage-affected region for repair. At each time step \( t \), the action space \( A_t = \{a^1_t, a^2_t, \dots, a^n_t \} \), where \( a^r_t \) corresponds to the features of the \( r \)-th region, and \( n \) is the number of candidate regions available for repair. Each action \( a^r_t \) is defined as \( a^r_t = \{ PI^r_t,  ST^r_t, CO^r \} \), where \( PI^r_t \) represents uncertainty information, specifically the prediction interval for repair duration in region \( r \). \( ST^r_t \) represents spatiotemporal information of region \( r \) at time \( t \). \( CO^r = \{ id^r, dis^r, sen^r \} \) denotes contextual information for region \( r \), including its identifier \( id^r \), distance \( dis^r \) from the repair team’s current location, and sensitive feature \( sen^r \).

    \item \textbf{Reward $\mathcal{R}$}: We assign the reward only at the end of each episode. Specifically, during all previous steps, the system provides no reward feedback (i.e., a reward of 0) until all regions have been restored with power. Once all regions are repaired, the system provides a reward reflecting the effectiveness of the entire repair process. In EPOPR, based on our optimization problem defined in Equation \ref{eq:1}, we use the negative average outage durations across all regions as the reward. The reward is calculated as follows: 
    \begin{equation} 
    R = - \left(\textstyle \sum_{r=1}^N T^r_{outage}\right) / N  \label{eq:13}
    \end{equation} 
    where $T^r_{outage}$ represents the outage duration for region $r$, and $N$ is the total number of power outage regions.
    \item \textbf{Cost $\mathcal{C}$}: We assign the cost only at the end of each episode, similar to our reward setting. The cost is defined as the maximum Wasserstein distance between the outage duration distributions of all possible pairs of sensitive groups. It is calculated as follows:
\begin{equation}
    C = \textstyle \max\limits_{k_1, k_2} W(P(T_{outage}|F_{k_1}), P(T_{outage}|F_{k_2}))
     \label{eq:14}
\end{equation}
where \( F_{k_1} \) and \( F_{k_2} \) represent any pair of sensitive groups. We require that the cost \( C \) does not exceed the predefined upper limit \( d \), which follows our equity criteria: ensure that the overall difference between the two groups remains within an acceptable range.
\end{itemize}

In the CMDP problem, the final optimization problem is:
\begin{equation}
    \max \: R \quad \text{s.t.} \quad C \leq d \label{eq:15}
\end{equation}
To address the aforementioned problem, we propose STA-SAC, which includes an Actor for action selection and four Critic networks for action value estimation. A key challenge in STA-SAC is effectively modeling the spatial-temporal correlations among these uncertainties while handling the dynamic nature of the action set. In the following section, we will introduce the Spatial-Temporal Attention Actor and show how it addresses this challenge.

\subsubsection{Spatial-Temporal Attentional Actor}
\begin{figure}[htbp]  \centering
    \centering
    \includegraphics[width=1\linewidth, keepaspectratio=true]{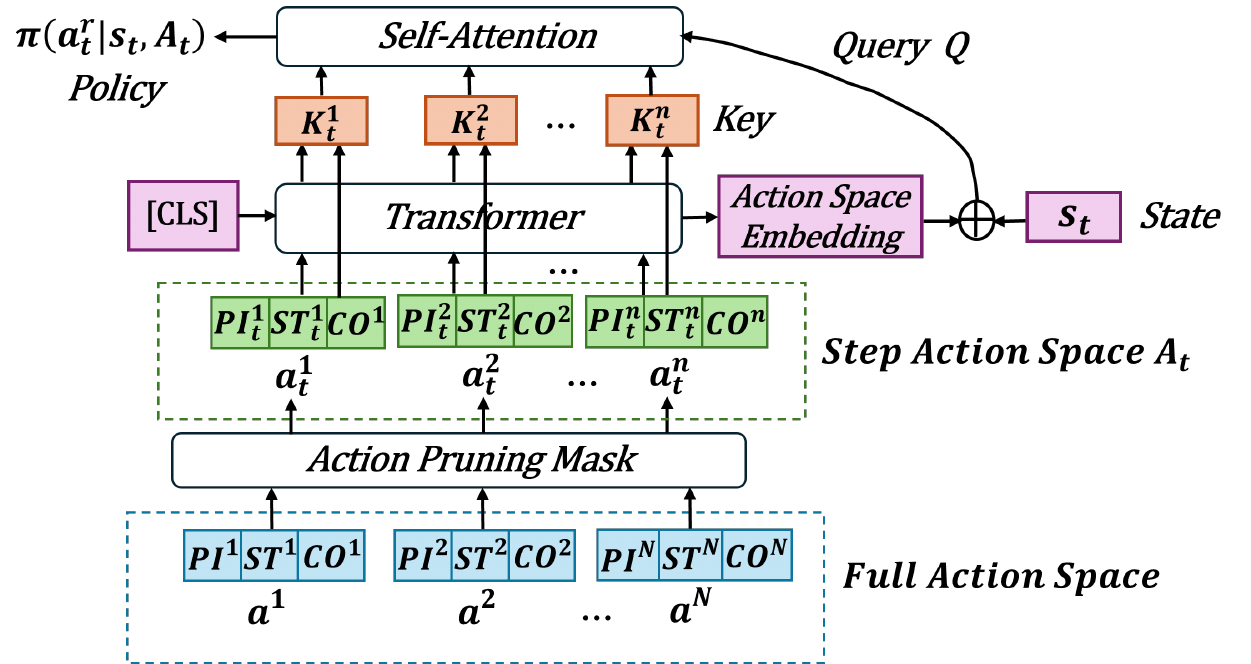}
    \caption{Spatial-Temporal Attentional Actor}
    \label{fig:framework}
\end{figure}

In reinforcement learning, the Actor selects the next action based on the agent’s current state, typically represented as \(\pi(a_t | s_t)\). Unlike traditional approaches, STA-SAC considers both the current state \(s_t\) and the entire action set \(A_t\), where each action \(a^r_t \in A_t\) is associated with uncertainty information \(PI^r_t\). Notably, the size of \(A_t\) is dynamic at each decision step in real-world settings. To address this, our Spatial-Temporal Attentional Actor leverages an Action Pruning Mask and a Transformer-based architecture~\cite{waswani2017attention} to handle the varying size of \(A_t\). A query-key attention mechanism is used to capture global dependencies within the action set and estimate the selection probability of each candidate action. The overall architecture of the Spatial-Temporal Attentional Actor is illustrated in Figure~\ref{fig:framework}.

In Spatial-Temporal Attention Actor, we assume that the total number of all outage regions is $N$, and we denote the set of these regions as the Full Action Space. Through the Action Prune Mask, we mask certain candidate actions to obtain the action space $A_t$ at step $t$. The masking logic mainly focuses on: excluding previously selected actions and actions whose corresponding regions are too far from the repair team.

After obtaining the Step Action Space $A_t$ at step $t$, we employ the Transformer to simultaneously perform two tasks: (1) embedding the entire action space $A_t$ and (2) integrating global information into each action. For the first task, we input the [CLS] token \cite{kenton2019bert} to obtain the embedding of the action space. For the second task, we concatenate the output from the Transformer for each action with the action itself to form the new action representation. 

Finally, we adopt the query-key mechanism from attention to generate our policy $\pi(a^r_t|s_t, A_t)$. We concatenate the Action Space Embedding with the agent's current state $s_t$ to form the Query $Q_t$, and treat all candidate actions as the Keys $[K^1_t, K^2_t, \dots, K^n_t]$. The action score $\chi^r_{t}$ for the $r$-th action $a^r_t$ is calculated as:$\chi^r_{t} = v_a \sigma(W_a [K^r_t \oplus Q_t])$, where $\oplus$ denotes concatenation. The policy $\pi(a^r_t|s_t, A_t)$ is then computed based on all candidate actions as:
\begin{equation}
    \pi(a^r_t|s_t, A_t) = \frac{\exp(\chi^r_{t})}{\sum_{i=1}^{n} \exp(\chi^i_{t})} \label{eq:16}
\end{equation}
Here, $\pi(a^r_t|s_t, A_t)$ represents the probability of selecting the action $a^r_t$ (i.e., the next region $r$ to be repaired) based on the current state $s_t$ and action space $A_t$ \cite{jiang2023faircod}.

\subsubsection{Training Method}
In STA-SAC, we adopt SAC-Lagrangian as the foundation of our RL model and incorporate the Lagrange multiplier $\lambda$ to integrate cost constraints into the reward-based optimization objective. STA-SAC consists of one Actor and four Critic networks. The Actor leverages our Spatial-Temporal Attention mechanism, while the Critic networks evaluate action values and guide decision-making. Specifically, the Actor is represented as $\pi_\theta(a^r_t|s_t, A_t)$, where $\theta$ denotes its learnable parameters. The Reward and Cost Critic networks are $Q_{w_{r_{\{1,2\}}}}(s_t, a_t)$ and $Q_{w_{c_{\{1,2\}}}}(s_t, a_t)$, respectively, with $w_{r_1}, w_{r_2}, w_{c_1}, w_{c_2}$ as their learnable parameters.

To update the reward Critic networks, we minimize the following loss function for $i = 1,2$:
\begin{align}
    L_{w_{r_i}} = & \small \frac{1}{MN} \sum_M \sum_{t=1}^N \Big( r_t + \gamma \min\limits_{j=1,2} Q_{w^-_{r_{j}}}(s_{t+1}, a_{t+1}) \notag \\ 
    & - \beta \log \pi_\theta(a^r_t|s_t, A_t) - Q_{w_{r_i}}(s_{t}, a_{t}) \Big) \label{eq:17}
\end{align}
Where \( M \) is the number of episodes per update cycle, \( N \) is the length of each episode, \( \gamma \) is the discount factor, and \( \beta \) determines entropy importance. \( w^-_{r_{j}} \) denotes the target network parameters of \( w_{r_{j}} \), a common RL technique. The cost Critic networks share a similar update strategy in Equation \ref{eq:17}.

\noindent To update the Actor, we minimize the following loss:
\begin{align}
        & L_\pi(\theta)  = \frac{1}{MN}\sum_M \sum_{t=1}^N \Big( \beta \log \pi_\theta(a^r_t|s_t, A_t) \notag  \\
        &\quad - \min\limits_{j=1,2} \left( Q_{w_{r_j}}(s_{t}, a_{t}) - \lambda Q_{w_{c_j}}(s_{t}, a_{t}) \right) \Big) \label{eq:18}
\end{align}

\noindent For Lagrange multiplier $\lambda$, it follows the updating rules:
\begin{align}
    \lambda \leftarrow \max\left( 0, \lambda - \eta_\lambda \nabla_{\lambda} 
    \Big( \min_{j=1,2} Q_{w_{c_j}}(s_{t}, a_{t}) - d \Big) \right) \label{eq:19}
\end{align}
where $\eta_\lambda$ denotes the step size for updating $\lambda$. More details of SAC-Lagrangian can be found in \cite{haarnoja2018soft}.

\section{Evaluation}

\subsection{Evaluation Methodology}
\noindent \textbf{Data:} We use two one-year power outage datasets in 2018 from Tallahassee, Florida, to evaluate the proposed predict-then-optimize power restoration framework. We can extract two key pieces of information from the data: power outage duration and repair duration for each region. We build a simulation environment based on these two datasets where the repair team travels from one region to another region according to the decisions made by algorithms. In our simulation, for every episode, the repair duration for each region is directly sampled from historical outage records, reflecting real-world repair duration patterns. The simulation ends when all regions are repaired, marking one complete episode.

\noindent \textbf{Metrics:}
We design two metrics to evaluate our approach: 

\textbf{Power Outage Duration \ensuremath{\hat{T}_{outage}}}: The average power outage duration across all regions. This metric quantifies the power restoration efficiency. The unit of this value is the hour.

\textbf{Power Restoration Inequity \ensuremath{WD_{inequity}}}: This metric quantifies disparity in outage durations across socio-economic groups. Specifically, we compute the Wasserstein distance between outage duration distributions of different sensitive groups and take the maximum value to capture the most significant inequality. The formal definition of the Wasserstein distance is in Equation \ref{eq:2}. Particularly, a smaller $WD_{inequity}$ indicates greater equity in power restoration.

\noindent \textbf{Baselines:} 
We compare EPOPR with the following baselines. (1) \textbf{GroundTruth (GT):} This baseline directly extracts the repair sequence from our real data. (2) \textbf{Greedy Method (GM):} The repair team always prioritizes repairing the region nearest to its current location. (3) \textbf{TSP-ST} \cite{tacs2016traveling}: We formulate the sequential repair task as a Traveling Salesman Problem (TSP) with service times, where each region has a repair duration. This approach minimizes the sum of repair and travel durations. (4) \textbf{RL-DTSP} \cite{zhang2021solving}: This method employs a DQN-based decision maker for route planning under deterministic predictions, where a Random Forest model provides each region’s repair duration. (5) \textbf{HRL-DPDP} \cite{ma2021hierarchical}: This method uses hierarchical reinforcement learning. The upper-level agent decides which type of region to repair first, while the lower-level agent selects the specific region. It relies on the same deterministic prediction inputs as RL-DTSP. (6) \textbf{ROPU} \cite{yan2024robust}: A predict-then-optimize approach with uncertainty consideration. It utilizes Conformal Prediction for prediction and an Actor-Critic framework for decision-making.

\subsection{Overall Performance}
We compare EPOPR with baselines and summarize the results in Table \ref{tab:Overall Performance}. In our experiment, we set the upper limit \( d \) to 8. The regional division method is based on the US Census Tract, and the region size $N$ in our experiment is set to be 55. Our key findings are:  
(1) Our method achieves the shortest average outage duration and the most equitable outage distribution. Specifically, it reduces outage duration by 20.44\% and restoration inequity by 49.87\% compared to the ground truth. Furthermore, compared to the best-performing baseline, ROPU, our method reduces these metrics by 3.60\% and 14.19\%, respectively.  
(2) Among baselines, the optimizer-based method TSP-ST outperforms RL-based RL-DTSP for small-scale optimization (where \( N \) is typically less than 100). The hierarchical RL-based HRL-DPDP performs poorly because multi-level objectives make the system prone to local optimization. Meanwhile, uncertainty-aware prediction-then-optimize methods, ROPU and EPOPR, significantly outperform others by quantifying uncertainty for reliable decisions.

\begin{table}[htbp]\footnotesize
    \centering 
    \renewcommand{\arraystretch}{1.15} 
    \resizebox{0.85\linewidth}{!}{  
    \begin{tabular}{lcc}
    \hline
    \textbf{Method} & $\hat{T}_{outage}\downarrow \; (h)$  & \ensuremath{WD_{inequity}} $\downarrow$  \\  \hline
    GT & 57.521 & 17.344 \\
    GM   & 53.287 & 16.365 \\
    TSP-ST & 49.337 &  11.389 \\
    RL-DTSP  & 50.853  &  11.952 \\
    HRL-DPDP & 56.924 &  15.725 \\
    ROPU &  47.471  &  10.132 \\
    \hline
    \textbf{EPOPR} & \textbf{45.762}  & \textbf{8.694} \\
    \hline
    \end{tabular}
    }
        \caption{Overall Performance. The $\downarrow$ indicates that a smaller value is better. The result is the mean value based on 10 experiments.}
    \label{tab:Overall Performance}
\end{table}

\subsection{Results for Repair Duration Prediction}
\begin{figure}[htbp]\centering
\begin{minipage}[htbp]{0.47\linewidth}
    \includegraphics[width = \linewidth,keepaspectratio=true]{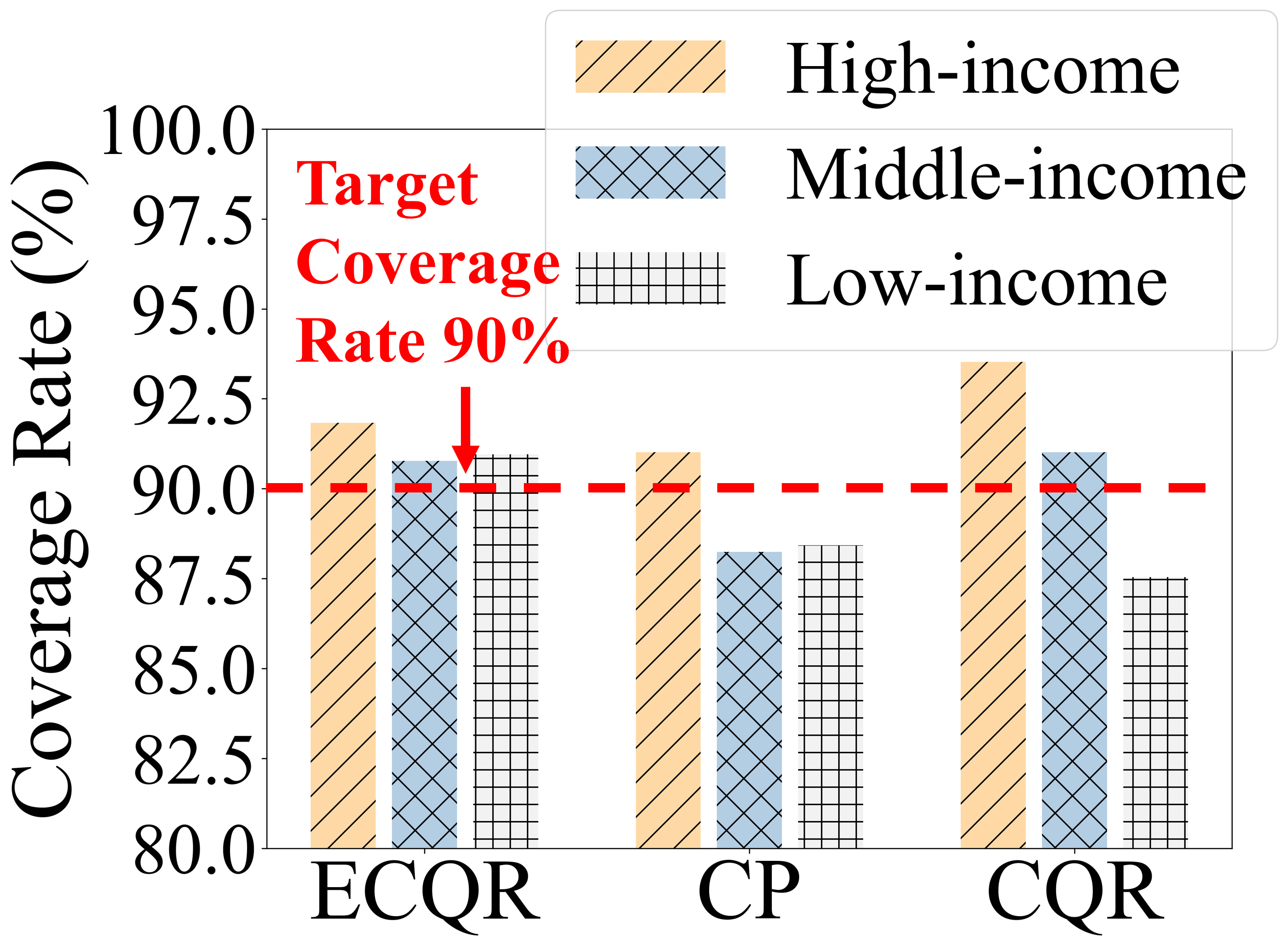}
    \captionsetup{font={small}}
    \caption{Prediction Interval Coverage Rate}
    \label{fig:Prediction Interval Coverage Rate}
\end{minipage}
\hspace{2mm}
\begin{minipage}[htbp]{0.45\linewidth}
    \includegraphics[width = \linewidth,keepaspectratio=true]{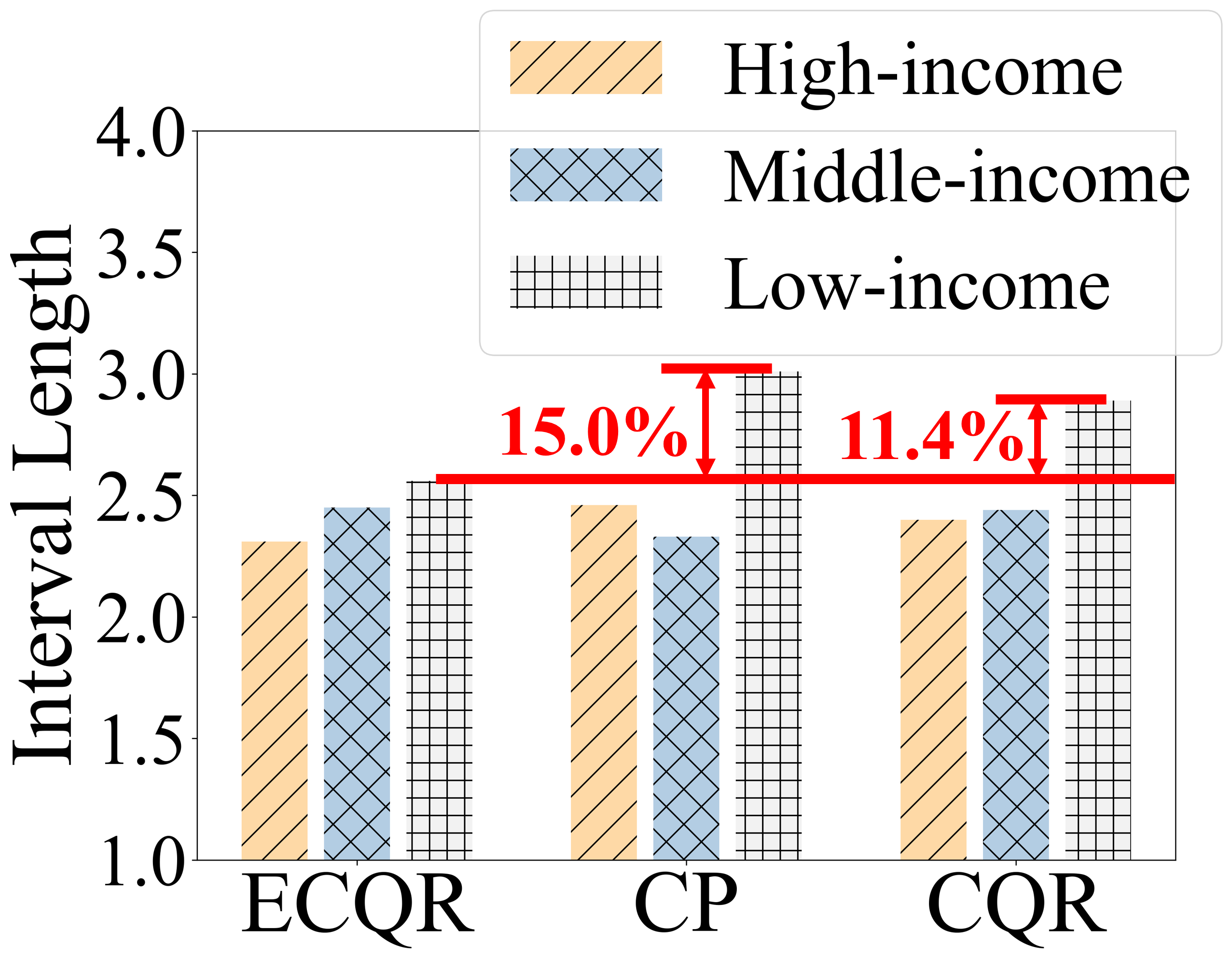}
    \captionsetup{font={small}}
    \caption{Prediction Interval Length}
    \label{fig:Prediction Interval Length}
\end{minipage}
\end{figure}

To assess the effectiveness of our uncertainty-aware outage prediction method, ECQR, we compare it with two commonly used baselines: Conformal Prediction (CP) \cite{shafer2008tutorial} and Conformalized Quantile Regression (CQR) \cite{romano2019conformalized}. Prediction interval quality is evaluated using coverage rate and interval length. Figure \ref{fig:Prediction Interval Coverage Rate} shows coverage rates for three sensitive groups, with a 90\% target (red dashed line). Compared to baselines, ECQR significantly reduces disparity among groups. While slightly lowering coverage in high-income regions compared to CQR, it notably improves coverage in low-income regions, ensuring all groups reach the target. Figure \ref{fig:Prediction Interval Length} shows that ECQR reduces interval length in low-income regions by 15.0\% and 11.4\% compared to the other two methods. Overall, ECQR delivers more reliable predictions for low-income regions, narrowing the gap with other groups for better equity.

\vspace{-10pt}

\subsection{Results for Outage Duration Distribution}
\begin{figure}[htbp]\centering
    \includegraphics[width=0.95\linewidth, keepaspectratio=true]{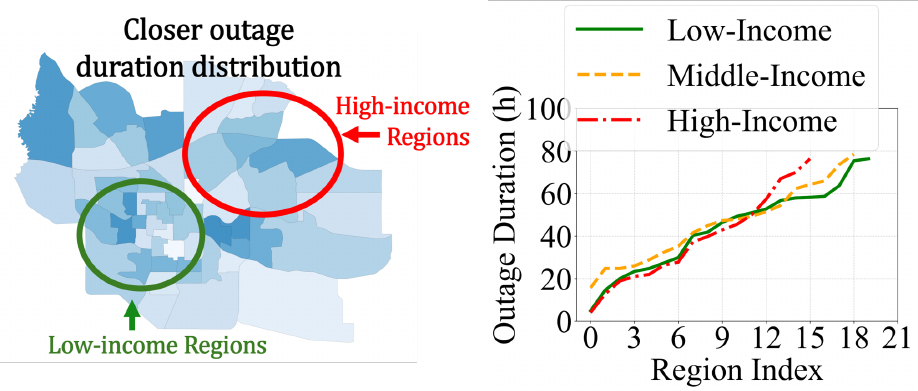}
    \caption{Outage Duration Distributions based on \textbf{EPOPR}}
    \label{fig:equal_dis}
\end{figure}

\begin{figure}[htbp]\centering
    \includegraphics[width=0.95\linewidth, keepaspectratio=true]{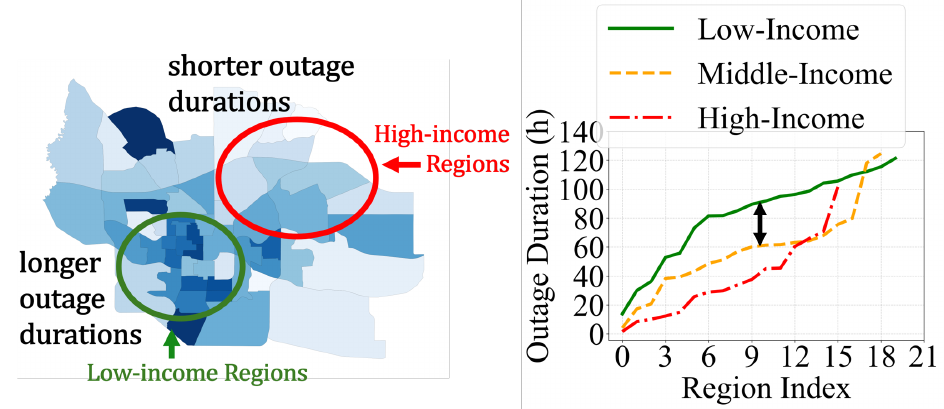}
    \caption{Outage Duration Distributions from Real Data}
    \label{fig:unequal_dis}
\end{figure}

We show the outage duration distribution under our method EPOPR and the current situation in Figure \ref{fig:equal_dis} and \ref{fig:unequal_dis}, respectively. The figures on the left depict power outage durations across the city, with darker blue indicating longer outages. The figures on the right present outrage durations of each region. By comparing the two figures, we can see our method reduces the differences in outage durations across sensitive groups, demonstrating its performance to improve equity.

\vspace{-10pt}

\section{Related Work}
\textbf{Uncertainty Quantification}: 
In recent years, uncertainty quantification (UQ) has attracted widespread attention in both prediction \cite{wen2023diffstg,rasul2021autoregressive,zhuang2022uncertainty} and decision-making \cite{yan2024robust,an2021uncertainty,ez2023reinforcement} tasks, as reliable uncertainty estimation not only enhances model interpretability but also provides a solid foundation for downstream decision tasks. Bayesian methods, such as Bayesian Neural Networks (BNN) \cite{gal2016dropout,lan2022scaling}, introduce probability distributions over the weights of the neural network and use Bayesian inference for uncertainty estimation. Predictive distribution methods explicitly model the output distribution to capture uncertainty. For example, Quantile Regression (QR) \cite{koenker1978regression,chung2021beyond} estimates various quantiles of the target distribution, yielding interval-based uncertainty estimates. In contrast, Deep Ensembles \cite{rahaman2021uncertainty} train multiple independent neural networks and aggregate their predictions to approximate the overall predictive uncertainty. Calibration methods aim to ensure that a model's predicted confidence levels accurately correspond to the true probabilities, thereby improving the reliability of uncertainty estimates. For example, temperature scaling \cite{kull2019beyond} adjusts the softmax outputs so that the predicted confidence scores more closely reflect the actual probability distribution.

\textbf{Post-Disaster Power Restoration}: Many studies have explored post-disaster power restoration from various disciplines. Here, we focus on data-driven approaches. \cite{afsharinejad2021large} employs unsupervised learning on real-world power grid data from New York and Massachusetts to analyze recovery capabilities under government policies. \cite{ji2016large,ganz2023socioeconomic} investigates the relationship between socioeconomic vulnerability and the differential impact of severe weather-induced power outages through large-scale data analysis. \cite{xu2020closing} investigates how the public service platform after disasters affects distributional equity in public service delivery, and how the government utilizes such a digital platform to improve post-disaster power restoration efficiency.

\section{Conclusion}
Motivated by our data-driven analysis, we found that the current power restoration decisions may be inequitable, as it is highly related to the number of repair requests from each region. This reliance on request volume creates significant disparities, often resulting in disadvantaged areas with fewer submissions receiving lower priority in the restoration process. To address this challenge, we design an equity-aware predict-then-optimize power restoration framework called EPOPR, which consists of two key components: (1) ECQR for repair duration prediction, ensuring equitable prediction intervals across all sensitive groups, and (2) STA-SAC for repair sequence decision-making, which aims to minimize the total outage duration while enforcing fairness constraints. Evaluations on real-world data demonstrate that EPOPR reduces the average outage duration by 3.60\% and mitigates regional inequities by 14.19\% compared to the best baseline.

\section*{Acknowledgments}
We sincerely thank all anonymous reviewers for their constructive comments. This work is partially supported by Florida State University, National Science Foundation under Grant Numbers 2315027 and 1940319.

\bibliographystyle{named}
\bibliography{ijcai25}

\end{document}